\documentclass[cameraready]{Interspeech}

\title{Zero-Shot Respiratory Sound Classification through LLM-Augmented Audio-Text Alignment}

\author[affiliation={1}]{Mustafa Talha}{İlerisoy}
\author[affiliation={2}]{Hung Manh}{Pham}
\author[affiliation={1}]{Mathias}{Funk}
\author[affiliation={1}]{Mykola}{Pechenizkiy}
\author[affiliation={1}]{Aaqib}{Saeed}

\address{
    $^1$ Eindhoven University of Technology, Eindhoven, Netherlands \\
    $^2$ Singapore Management University, Singapore
}
\email{\{m.t.ilerisoy, m.funk, m.pechenizkiy, a.saeed\}@tue.nl, hm.pham.2023@phdcs.smu.edu.sg, Github: \url{https://github.com/mtilerisoy/REACH}}


\usepackage{comment}
\usepackage{multirow}
\usepackage{array}
\usepackage{booktabs}
\usepackage{etoolbox}
\usepackage{soul}      
\usepackage[table]{xcolor}
\usepackage{xspace}
\usepackage{relsize}
\usepackage{amsmath}
\usepackage{amssymb} 
\usepackage{pifont} 
\usepackage{booktabs, multirow, colortbl, xcolor}
\definecolor{shadegray}{gray}{0.92}
\DeclareMathOperator*{\argmax}{arg\,max}

\colorlet{shadegray}{gray!10}

\newbool{showComments}
\setbool{showComments}{true}  

\ifbool{showComments}{

}

\begin{document}

\maketitle

\begin{abstract}
Self-supervised respiratory encoders lack semantic grounding in clinical domain needed for zero-shot inference, limiting their utility without task-specific labeled data. We propose a framework that aligns these encoders with medical terminology in a shared latent space turning them into a zero-shot-capable foundation model. To address paired data scarcity, we use a medical LLM to synthesize structured reports from metadata, creating dense semantic anchors for contrastive learning. Our training combines a sigmoid-based contrastive loss with encoder's native SSL objective and similarity-aware negative sampling to sharpen pathological boundaries. Across 9 tasks on 6 datasets, our method achieves a 61.3\% mean zero-shot AUC, surpassing CLAP (51.4\%) and Qwen2-Audio (54.9\%) while reaching the highest linear probing AUC (71.6\%) with only 43\% of data used by full-scale baselines, showing that structured semantic alignment outperforms large-scale, general-purpose models in clinical diagnostics.

\end{abstract}

\section{Introduction}
Respiratory auscultation remains one of the most widely practiced diagnostic procedures worldwide \cite{Cook2022Body}, yet its interpretation is highly subjective and dependent on clinical expertise that is unevenly distributed~\cite{OPERA50_Reichert2008}. AI-driven auscultation tools offer a path toward scalable, objective screening~\cite{DOHENY2023Respiratory, Moberg2025Lung, Drummond2024Challenges}, particularly in low-resource settings where pulmonologists are unavailable~\cite{ZERO_Liu2025Multimodal, Rare_Hasani2022Artificial}. For such tools to be clinically viable, however, they must generalize to novel pathologies \emph{without} task-specific labeled data, a capability known as zero-shot inference, which remains largely unattained in respiratory audio modeling \cite{Rath2025Tuberculosis}.

Self-supervised learning has produced powerful acoustic encoders that form the backbone of current respiratory AI. General-purpose models~\cite{AudioMAEHuang2023, VGGishHershey2017} learn spectro-temporal representations through masked autoencoding on broad audio corpora, while domain-specific efforts~\cite{OPERAZhang2024} tailor these to respiratory events via large-scale pre-training. These encoders function as \emph{acoustic experts}, resolving fine-grained auscultatory patterns (crackles, wheezes, stridor) with high fidelity. Yet, their embedding spaces remain \emph{semantically opaque}: clusters corresponding to distinct pathologies may be linearly separable, but neither is anchored to the medical concept it encodes. Consequently, every new clinical task requires supervised fine-tuning with scarce, expert-curated annotations~\cite{Rare_Berger2024TacklingDS}, creating a bottleneck that limits real-world deployment.

Multimodal contrastive learning addresses this by aligning audio and text in a shared space, enabling zero-shot classification via natural language prompts that are validated for vision~\cite{CLIPradford2021learning} and general audio~\cite{CLAPElizalde2023}. However, on clinical respiratory benchmarks, general-purpose audio-text models fail to outperform unimodal baselines~\cite{OPERAZhang2024, Niizumi2025_Interspeech}: they are grounded in everyday captions and lack clinical language. Training a medical audio-text model from scratch would require paired audio-report datasets that do not exist at scale for respiratory sounds~\cite{Shokouhmand2025Scarcity}.

We observe that strong acoustic features and clinical language understanding already exist in isolation: respiratory encoders~\cite{OPERAZhang2024} capture the diagnostic signal, while medical text encoders~\cite{sellergren2025medgemma} capture disease-level semantics. The core problem is not representation learning but \emph{representation alignment} \cite{Nakada2023Understanding}. While other methods align vision-language spaces \cite{yang2025languageimagealignmentfixedtext} or employ generative instruction-tuning for respiratory health \cite{pmlr-v259-zhang25a}, we address paired-data scarcity through sample-efficient contrastive alignment. Building on this, we propose  REACH (REport-Augmented Contrastive alignment for respiratory Health), a \textbf{semantic alignment framework} that re purposes pre-trained unimodal respiratory encoders into zero-shot capable multimodal tools through targeted post-training. Our approach treats existing encoders as complementary assets and focuses on learning the bridge between them. The design is \emph{modular}, accepting any transformer-based audio backbone; \emph{data efficient}, requiring no paired audio report corpora; and \emph{non-destructive}, preserving acoustic fidelity throughout.

\begin{figure*}[t]
  \centering
  \vspace{-10mm}
  \includegraphics[width=1.0\textwidth]{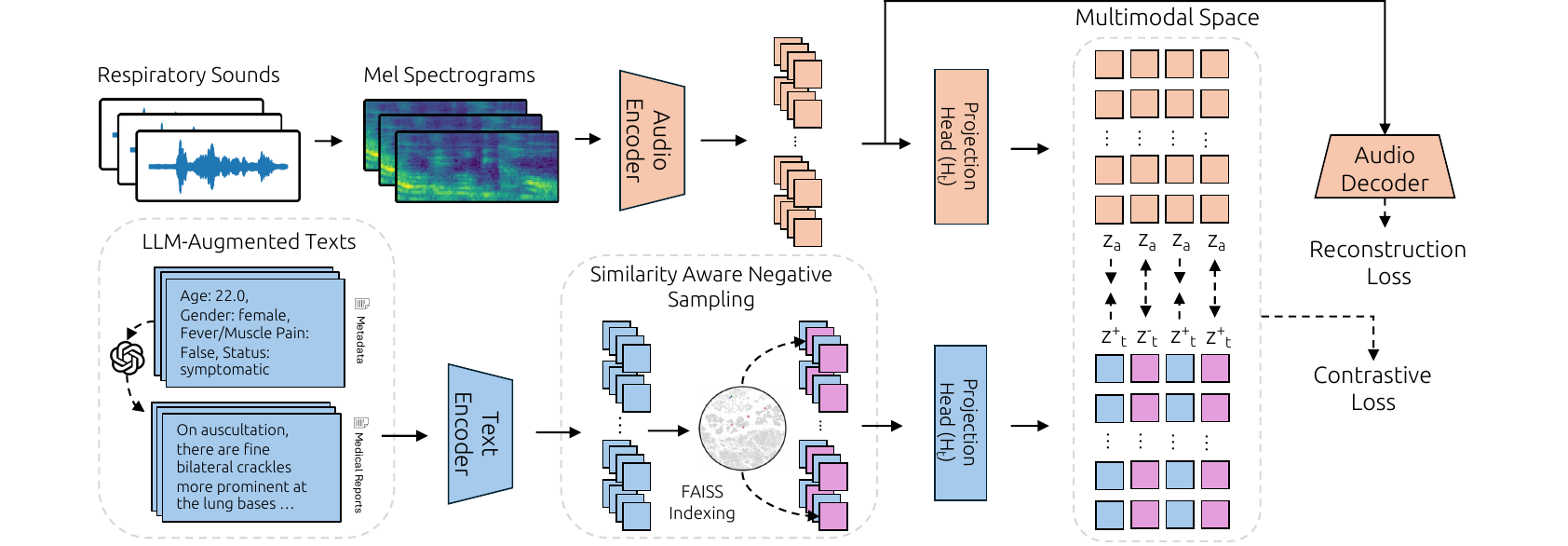}
  \caption{Overview of REACH: a semantic alignment framework that repurposes pre-trained unimodal encoders for zero-shot respiratory sound classification via LLM-augmented report synthesis, similarity-aware negative sampling, and structure-preserving contrastive alignment.}
  \label{fig:System Overview}
  \vspace{-5mm}
\end{figure*}

Our framework addresses three challenges: (1) absent paired data; we use an off-the-shelf medical-grade LLM~\cite{openai2024gpt4technicalreport} to synthesize reports from metadata, creating semantic anchors for contrastive learning; (2) low textual diversity; we employ FAISS-based~\cite{FAISS} similarity-aware negative sampling to mine distant negatives; and (3) catastrophic feature degradation; we combine a sigmoid contrastive loss~\cite{SigLIP} with the encoder's native reconstruction objective as a structural regularizer. The text encoder~\cite{sellergren2025medgemma} remains frozen as a fixed semantic reference. Across 9 tasks on 6 datasets, our method achieves 61.3\% mean zero-shot AUC, surpassing CLAP (51.4\%) and Qwen2-Audio (54.9\%), while attaining the highest linear probing AUC (71.6\%) with only 43\% of the training data used by the full-scale baseline~\cite{OPERAZhang2024}.

The text branch employs a medical text encoder~\cite{sellergren2025medgemma} that remains fully frozen during alignment as a fixed semantic reference; only the audio branch and projection heads are optimized. We evaluate on 9 tasks from 6 publicly available respiratory datasets spanning in-domain and out-of-domain settings. Our method achieves a mean zero-shot AUC of 61.3\%, surpassing CLAP (51.4\%) and the 7B parameter Qwen2 Audio (54.9\%), while using only 43\% of the pre-training data available to the full-scale baseline~\cite{OPERAZhang2024}. Importantly, alignment does not compromise unimodal capability: our model attains the highest mean linear probing AUC (71.6\%), exceeding even the baseline trained on the complete proprietary corpus. Our contributions are as follows:
\begin{enumerate}[leftmargin=*, itemsep=2pt, topsep=2pt]
    \item A \textbf{modular semantic alignment framework} (REACH) that transforms pre-trained unimodal respiratory encoders into zero-shot capable multimodal models, decoupling acoustic pre-training from clinical language grounding.
    \item An \textbf{LLM augmented report synthesis} pipeline that converts discrete metadata into clinically structured text, creating semantic anchors that eliminate the need for paired audio report datasets.
    \item A \textbf{structure-preserving alignment strategy} that reconciles cross-modal transfer with preservation of acoustic features by jointly optimizing a sigmoid contrastive objective~\cite{SigLIP} with the encoder's native reconstruction loss, complemented by similarity-aware negative sampling.
    \item \textbf{Comprehensive empirical validation} across 9 tasks on 6 datasets, demonstrating that targeted alignment with 57\% less data outperforms both full-scale unimodal pre-training and general-purpose audio language models of significantly larger capacity.
\end{enumerate}

\vspace{-0.1cm}
\section{Methodology}
\subsection{Problem Formulation}

Let $f_a$ denote a pre-trained unimodal audio encoder and $f_t$ a pre-trained text encoder, producing embeddings $\mathbf{x}_a \in \mathbb{R}^{m}$ and $\mathbf{x}_t \in \mathbb{R}^{n}$, respectively. These embedding spaces are independently learned and share no semantic correspondence. Our objective is to learn projection heads $H_a: \mathbb{R}^{m} \rightarrow \mathbb{R}^{d}$ and $H_t: \mathbb{R}^{n} \rightarrow \mathbb{R}^{d}$ that map both modalities into a shared space $\mathbb{R}^{d}$ where semantically corresponding pairs are aligned.

Let $\mathbf{z}_a = H_a(\mathbf{x}_a)$ and $\mathbf{z}_t^{+} = H_t(\mathbf{x}_t^{+})$ denote projections of a matched pair and $\mathbf{z}_t^{-} = H_t(\mathbf{x}_t^{-})$ a non-corresponding report. We seek projections satisfying:
\begin{equation}
    S(\mathbf{z}_a,\, \mathbf{z}_t^{+}) > S(\mathbf{z}_a,\, \mathbf{z}_t^{-}) \quad \forall\, \mathbf{x}_t^{-} \neq \mathbf{x}_t^{+}
\end{equation}
where $S(\cdot, \cdot)$ denotes cosine similarity. For instance, the projection of a recording with prominent wheezes should be closer to a report describing \textit{``expiratory wheezes in a 70-year-old female''} than to one describing \textit{``clear breath sounds with no adventitious findings.''}
Once aligned, zero shot classification reduces to $\argmax_c\, S(\mathbf{z}_a, \mathbf{z}_{t_c})$ over class specific text anchors $\{\mathbf{z}_{t_c}\}_{c=1}^{C}$, each generated by encoding a clinical prompt for class $c$.

\subsection{Semantic Anchor Generation via LLM Augmented Report Synthesis}

Contrastive alignment requires semantically rich, paired text for each audio sample. Existing respiratory datasets provide only discrete metadata: sound type (e.g., cough, breathing cycle), adventitious sound labels (e.g., wheeze, crackle), recording location (e.g., posterior lower lobe), patient demographics, and disease diagnosis. These categorical fields lack the narrative structure needed to serve as effective linguistic anchors.

We employ a medical-grade off-the-shelf LLM (i.e., GPT-4~\cite{openai2024gpt4technicalreport}) to transform this metadata into standardized clinical reports, conditioned on a prompt instructing the model to adopt the role of a pulmonologist:

\vspace{4pt}
\noindent\fbox{\parbox{0.95\columnwidth}{\small\textit{\scriptsize{You are a Pulmonologist tasked with interpreting respiratory auscultation findings. Based on the given conditions, write 2--3 lines covering clinically relevant information. Only use the information given to write about conditions. Do NOT mention anything about further evaluation or characterization.}}}}
\vspace{4pt}

This prompt enforces two constraints: restricting the LLM to the provided metadata prevents hallucination of ungrounded clinical details, and prohibiting follow-up recommendations keeps the anchors tightly coupled to the observable acoustic content. To prevent information leakage, the metadata used during alignment is strictly partitioned from evaluation data; at inference, the model encounters clinical concepts (e.g., \textit{COPD}) never paired with audio during training.

\vspace{-0.2cm}
\subsection{Model Architecture}

Our framework is \emph{architecturally modular}, accepting any pair of transformer-based unimodal encoders. We instantiate it with two backbones selected for complementary strengths.

\vspace{3pt}
\noindent\textbf{Audio Encoder.}\quad We adopt a respiratory-specific transformer~\cite{OPERAZhang2024} pre-trained via masked spectrogram reconstruction on a large-scale mel spectrogram corpus (see Table~\ref{tab:results} for the specific variant), representing the current state of the art for respiratory feature extraction.

\vspace{3pt}
\noindent\textbf{Text Encoder.}\quad We employ the text branch of MedSigLIP~\cite{sellergren2025medgemma}, a medical vision language encoder pre-trained on paired chest radiographs and clinical reports via sigmoid contrastive loss. Its representations encode clinical nomenclature and report structure, making it an ideal fixed semantic reference. The vision encoder is discarded; only the text encoder is retained. Notably, this encoder has never seen respiratory audio, so all cross-modal alignment is learned through our framework.

\vspace{3pt}
\noindent\textbf{Projection Heads.}\quad We introduce lightweight projection heads $H_a$ and $H_t$, each consisting of a linear layer followed by layer normalization, mapping $m$ and $n$ dimensional features into the shared $d$ dimensional space. 

\vspace{-0.15cm}
\subsection{Alignment Objective}

We optimize a dual objective that drives cross-modal alignment while preserving the encoder's pre-trained feature geometry:
\begin{equation}
    \mathcal{L}_{\text{total}} = \mathcal{L}_{\text{contrastive}} + \mathcal{L}_{\text{MSE}}
\end{equation}

\vspace{3pt}
\noindent\textbf{Contrastive Alignment ($\mathcal{L}_{\text{contrastive}}$).}\quad We adopt a SigLIP-based~\cite{SigLIP} formulation that treats each audio text pair as an independent binary classification problem. For a batch of $N$ pairs:
\begin{equation}
    \mathcal{L}_{\text{contrastive}} = -\frac{1}{N^2}\sum_{i=1}^{N}\sum_{j=1}^{N} \log \sigma\left(y_{ij}\left(\alpha \cdot S(\mathbf{z}_a^i, \mathbf{z}_t^j)\right)\right)
\end{equation}
where $y_{ij} = +1$ if $(i,j)$ is a matched pair and $y_{ij} = -1$ otherwise, $\sigma$ is the sigmoid function, and $\alpha$ is a learnable temperature parameter. Unlike softmax-based InfoNCE~\cite{CLIPradford2021learning}, this formulation requires no global batch normalization, making it robust to the small batch sizes typical in medical settings.

\vspace{3pt}
\noindent\textbf{Masked Reconstruction Regularizer ($\mathcal{L}_{\text{MSE}}$).}\quad Contrastive fine tuning alone risks distorting the pre-trained acoustic manifold \cite{Kumar2022Fine}. We retain the encoder's original self-supervised objective: during each forward pass, random spectrogram patches are masked, and the encoder reconstructs them. This MSE loss acts as a \emph{structural regularizer}, anchoring representations to their pre-trained geometry while the contrastive term reshapes the space for cross-modal compatibility.

\vspace{-0.15cm}
\subsection{Similarity Aware Negative Sampling}

In clinical domains, semantically distinct conditions can produce textually similar reports (e.g., ``wheezes in a 65-year-old male with asthma'' vs.\ ``wheezes in a 60-year-old male with COPD''), rendering in batch random negatives insufficiently contrastive.

\vspace{3pt}
\noindent\textbf{Offline Indexing.}\quad Before training, we encode the entire report corpus with the frozen text encoder and construct a FAISS index~\cite{FAISS} over the resulting embeddings.

\vspace{3pt}
\noindent\textbf{Online Negative Swapping.}\quad During each training step, 50\% of audio samples in the batch undergo negative swapping: we query the FAISS index to retrieve the $k$\textsuperscript{th} furthest embedding ($k{=}10$) from the positive anchor, which replaces the in-batch negative. Selecting $k{=}10$ rather than the absolute furthest point avoids degenerate outlier negatives while ensuring substantial semantic distance. This forces the model to maximize the margin between unrelated clinical pathologies in $\mathbb{R}^{d}$.

\vspace{-0.15cm}
\subsection{Training Strategy}

The text encoder remains fully frozen; only the audio encoder and projection heads are updated (lr = $1 \times 10^{-5}$, $100$ epochs). This asymmetry ensures the acoustic manifold migrates toward the stable clinical text space, and that inference prompts are interpreted in the same semantic frame used during training.

\vspace{-0.1cm}
\section{Experimental Setup}
\subsection{Evaluation Benchmark}
We evaluate on a comprehensive benchmark \cite{OPERAZhang2024} comprising 6 publicly available respiratory sound datasets to evaluate representation quality, generalization, and cross modal alignment. The benchmark consists of 9 tasks grouped into 5 \emph{in domain} (ID) tasks, drawn from datasets used during pre training and alignment, and 4 \emph{out of domain} (OOD) tasks, drawn from datasets the model has never encountered in any training phase. All tasks are binary classification problems except T9, which is a five class COPD severity grading task. Table~\ref{tab:dataset_details} summarizes the dataset characteristics, task definitions, and class distributions.

 \begin{table}[t]
    \vspace{-5pt}
    \centering
    \caption{Summary of curated datasets; shaded rows denote OOD datasets withheld from pre-training and alignment.}
    \label{tab:dataset_details}
    \vspace{-5pt}
    \resizebox{1.0\columnwidth}{!}{%
    \begin{tabular}{rllll}
    \toprule
    Dataset & ID & Task & Modality & \#Samples \\
    \midrule
    CovidUK~\cite{covidUK}      & T1 & Covid/Non-covid      & Exhalation & 840 / 1660 \\
                                & T2 & Covid/Non-covid      & Cough & 840 / 1660 \\
    COUGHVID~\cite{COUGHVID}    & T3 & Covid/Non-covid      & Cough & 547 / 5628 \\
                                & T4 & Female/Male          & Cough & 2468 / 4795 \\
    ICBHI~\cite{ICBHI2017}      & T5 & COPD/Healthy         & Lung sounds & 793 / 35 \\

    \rowcolor{shadegray}
    Coswara~\cite{Coswara}      & T6 & Smoker/Non-smoker    & Cough & 201 / 747 \\
    \rowcolor{shadegray}
                                & T7 & Female/Male          & Cough & 759 / 1737 \\
    \rowcolor{shadegray}
    KAUH~\cite{KAUH}            & T8& Obstructive/Healthy  & Lung sounds & 129 / 105 \\
    \rowcolor{shadegray}
    Resp.@TR~\cite{Resp_at_tr}  & T9& 5-class COPD severity        & Lung sounds & 72/60/84/84/204 \\
    \bottomrule
    \end{tabular}
    }
    \vspace{-7mm}
\end{table}
 
\vspace{-0.15cm}
\subsection{Evaluation Protocol}

We evaluate along three axes. Linear probing: a frozen encoder plus single linear layer (5-seed average). kNN: non-parametric classification measuring embedding geometry directly. Zero-shot: each class is represented by a clinically descriptive text prompt encoded into the shared space; classification assigns each sample to the nearest text anchor by cosine similarity. We report AUC (\%) throughout.

\vspace{-0.15cm}
\subsection{Baselines}

We organize baselines into three categories to contextualize our results against models with fundamentally different capabilities:

\noindent\textbf{Unimodal encoders} (linear probing and kNN only): OpenSMILE~\cite{opensmile}, VGGish~\cite{VGGishHershey2017}, AudioMAE~\cite{AudioMAEHuang2023}, and the OPERA family~\cite{OPERAZhang2024} (OCT, OCE, OGT). \textbf{General-purpose audio-text}: CLAP~\cite{CLAPElizalde2023} (all three protocols). \textbf{Audio-language decoders} (zero-shot only): Qwen2-Audio 7B~\cite{qwen} and Audio-Flamingo-3~\cite{flamingo}, included to test whether scale compensates for domain alignment.

\vspace{-0.15cm}
\subsection{Data Fairness and Training Conditions}

A critical consideration in our evaluation is data parity. The full OPERA training corpus contains datasets with varying access conditions: a subset is publicly available, while the remainder requires institutional data use agreements. To ensure a reproducible and fair comparison, we retrain OPERA variants (denoted $\dagger$ in Table~\ref{tab:results}) using only the openly accessible subset, comprising approximately 43\% of the original training volume. Our alignment framework operates on this same subset, ensuring all comparisons against $\dagger$ variants are data-matched and isolate the effect of our alignment strategy from differences in training scale. We additionally report full-corpus OPERA results (without $\dagger$) to contextualize absolute performance, but exclude these from the primary ranking due to the differing training data.

\vspace{-0.1cm}
\section{Results}
\begin{table}[t]
    \centering
    \setlength{\tabcolsep}{2.5pt}
    \renewcommand{\arraystretch}{0.88}
    \scriptsize
    \caption{Benchmark results (\% AUC). $^\dagger$: trained on open access data only ($\approx$43\%). \colorbox{shadegray}{Shaded}: full corpus, excluded from ranking. $\Delta$: gain over OGT$^\dagger$. \textbf{Bold} and \underline{underline} indicates the best and the second best respectively.}
    \label{tab:results}
    \vspace{-3pt}
    \resizebox{\columnwidth}{!}{%
    \begin{tabular}{l ccccc cccc c}
        \toprule
        & \multicolumn{5}{c}{\footnotesize In Domain} & \multicolumn{4}{c}{\footnotesize OOD} & \\
        \cmidrule(lr){2-6} \cmidrule(lr){7-10}
        & T1 & T2 & T3 & T4 & T5 & T6 & T7 & T8 & T9 & Avg \\
        
        \midrule
        \multicolumn{11}{l}{\textit{Linear Probing} $\uparrow$} \\[-1pt]
        \noalign{\vskip 2pt}
        CLAP & 56.5 & 64.8 & \textbf{59.9} & 66.5 & \textbf{93.3} & \textbf{68.0} & 74.2 & \textbf{69.7} & 63.6 & 68.5 \\
        OpenSMILE & 55.0 & 64.9 & 53.7 & 67.7 & 57.9 & 53.4 & 75.3 & 63.6 & 49.4 & 60.1 \\
        VGGish & 58.0 & 55.7 & 53.8 & 60.0 & 60.5 & 50.7 & 60.6 & 60.5 & 59.0 & 57.6 \\
        AudioMAE & 54.9 & 61.6 & 55.4 & 62.8 & 88.6 & 54.9 & 72.4 & 61.6 & 51.0 & 62.6 \\
        \rowcolor{shadegray} OCT & 58.6 & 70.1 & 57.8 & 79.5 & 85.5 & 68.5 & 87.4 & 72.2 & 62.5 & 71.3 \\
        \rowcolor{shadegray} OCE & 55.1 & 62.9 & 56.6 & 72.1 & 87.2 & 67.4 & 80.1 & 74.1 & 68.3 & 69.3 \\
        \rowcolor{shadegray} OGT & 60.5 & 67.7 & 55.2 & 73.5 & 74.1 & 65.0 & 82.5 & 70.3 & 60.6 & 67.7 \\
        OCT$^\dagger$ & 58.7 & 67.2 & \underline{56.7} & 76.9 & 88.0 & \underline{67.4} & \underline{84.4} & \underline{68.4} & \underline{66.2} & \underline{70.4} \\
        OCE$^\dagger$ & 55.1 & 65.4 & 54.8 & 70.6 & 87.1 & 64.2 & 75.0 & 63.0 & 65.2 & 66.7 \\
        OGT$^\dagger$ & 56.2 & 64.7 & 55.6 & 70.3 & 74.5 & 62.9 & 77.8 & 64.7 & 61.7 & 65.4 \\ \midrule
        \textbf{REACH (Ours)} & \textbf{63.8} & \textbf{69.8} & 56.3 & \textbf{78.4} & \underline{90.2} & 66.1 & \textbf{86.3} & 66.9 & \textbf{66.6} & \textbf{71.6} \\
        $\Delta$ & {\scriptsize 7.6} & {\scriptsize 5.1} & {\scriptsize 0.7} & {\scriptsize 8.1} & {\scriptsize 15.7} & {\scriptsize 3.2} & {\scriptsize 8.5} & {\scriptsize 2.2} & {\scriptsize 4.9} & {\scriptsize 6.2} \\

        \midrule
        \multicolumn{11}{l}{\textit{$k$\text{-}NN} $\uparrow$} \\[-1pt]
        \noalign{\vskip 2pt}
        CLAP & \underline{54.4} & 57.9 & 51.8 & 59.0 & 89.0 & 53.3 & 61.4 & \textbf{71.5} & \underline{57.3} & \underline{61.7} \\
        OpenSMILE & 49.7 & 51.3 & 47.1 & 52.2 & 77.7 & 52.2 & 51.3 & \underline{71.3} & 55.2 & 56.4 \\
        VGGish & 53.2 & 53.9 & 50.3 & 56.8 & 81.6 & 51.3 & 61.3 & 53.9 & 47.3 & 56.6 \\
        AudioMAE & 52.4 & 56.5 & 48.4 & 52.2 & \underline{89.5} & 51.6 & 58.0 & 58.7 & 53.1 & 57.8 \\
        \rowcolor{shadegray} OCT & 48.7 & 60.9 & 55.2 & 71.0 & 62.7 & 65.8 & 75.9 & 66.6 & 55.4 & 62.5 \\
        \rowcolor{shadegray} OCE & 49.3 & 53.8 & 51.7 & 60.6 & 67.4 & 56.2 & 72.1 & 65.8 & 57.4 & 59.3 \\
        \rowcolor{shadegray} OGT & 51.3 & 55.6 & 50.9 & 60.7 & 66.0 & 57.1 & 68.5 & 59.4 & 49.0 & 57.6 \\
        OCT$^\dagger$ & 48.6 & \underline{59.3} & \textbf{56.7} & \textbf{68.8} & 74.4 & \underline{61.5} & \underline{72.9} & 57.0 & 52.4 & 61.3 \\
        OCE$^\dagger$ & 48.9 & 56.2 & 50.0 & 62.1 & 73.3 & 57.5 & 67.6 & 57.1 & 54.7 & 58.6 \\
        OGT$^\dagger$ & 50.1 & 56.5 & 52.3 & 61.4 & 68.3 & 56.1 & 72.4 & 58.3 & 53.8 & 58.8 \\ \midrule
        \textbf{REACH (Ours)} & \textbf{55.5} & \textbf{62.4} & \underline{52.4} & \underline{67.2} & \textbf{90.1} & \textbf{64.4} & \textbf{77.0} & 56.9 & \textbf{60.6} & \textbf{65.2} \\
        $\Delta$ & {\scriptsize 5.6} & {\scriptsize 5.9} & {\scriptsize 0.1} & {\scriptsize 5.8} & {\scriptsize 21.8} & {\scriptsize 8.3} & {\scriptsize 6.4} & {\scriptsize $-$1.4} & {\scriptsize 6.8} & {\scriptsize 6.4} \\

        \midrule
        \multicolumn{11}{l}{\textit{Zero Shot} $\uparrow$} \\[-1pt]
        \noalign{\vskip 2pt}
        CLAP & 52.4 & \textbf{62.6} & 43.6 & 63.4 & 51.8 & 48.6 & 61.5 & 31.3 & 47.5 & 51.4 \\
        Qwen2 Audio 7B & 52.0 & 55.0 & 47.8 & 63.1 & 75.0 & 46.7 & 56.8 & 46.9 & 51.2 & 54.9 \\
        Flamingo 3 & 48.7 & 53.7 & 51.1 & 68.9 & 25.8 & 53.4 & \textbf{71.8} & 42.9 & \textbf{65.1} & 53.5 \\ \midrule
        \textbf{REACH (Ours)} & \textbf{58.7} & \underline{61.5} & \textbf{51.3} & \textbf{67.1} & \textbf{76.7} & \textbf{67.4} & \underline{65.0} & \textbf{52.0} & \underline{52.1} & \textbf{61.3} \\
        \bottomrule
    \end{tabular}
    }
    \vspace{-0.6cm}
\end{table}

Table~\ref{tab:results} presents the full benchmark results. We analyze them along three axes: preservation of unimodal capability, improvement in latent space structure, and zero-shot cross-modal transfer.

\vspace{3pt}
\noindent\textbf{Alignment enhances unimodal features, rather than degrading them.}\quad Restricting the acoustic encoder~\cite{OPERAZhang2024} to the openly accessible 43\% of its original training corpus incurs a 2.3 point drop in mean linear probing AUC (67.7 $\rightarrow$ 65.4). Our alignment framework not only recovers this loss but also surpasses the full corpus baseline, reaching a 71.6 mean AUC (+6.2 over OGT$^\dagger$, +3.9 over the original OGT). Gains span all tasks in comparison with OGT$^\dagger$, with the largest improvements on T5 (+15.7), T7 (+8.5), T4 (+8.1), and T1 (+7.6). This indicates that the semantic anchors introduced during alignment provide a complementary training signal that enriches acoustic representations beyond what additional audio data alone achieves.

\vspace{3pt}
\noindent\textbf{Contrastive alignment reorganizes the audio manifold into clinical clusters.}\quad The $k$NN protocol measures embedding geometry without learned classification parameters. Our method achieves a 65.2 mean AUC, outperforming all baselines, including OGT$^\dagger$ (58.8) by 6.4 points. The strongest gain appears on T5 (COPD/Healthy: 90.1 vs.\ 68.3), where medical text anchors provide the semantic scaffolding to separate pathological from healthy lung sounds. The only regression is T8 ($-$1.4), which we attribute to its small sample size (234 samples), insufficient to form stable neighborhoods in the projected space.

\vspace{3pt}
\noindent\textbf{Targeted medical alignment outperforms both general-purpose and large-scale models at zero-shot inference.}\quad The zero-shot setting is the centerpiece of our evaluation: the model must classify audio solely via cosine similarity to text anchors, with no labeled data. OPERA baselines~\cite{OPERAZhang2024} cannot operate here due to their unimodal nature. Our method achieves 61.3\% mean AUC, substantially exceeding CLAP (51.4\%) and Qwen2 Audio 7B (54.9\%). CLAP's failure despite being designed for zero-shot audio classification confirms that general-purpose text grounding is insufficient for clinical domains. That a 7B parameter generative model also falls short demonstrates that scale alone cannot substitute for domain-specific semantic alignment. Among individual tasks, strong results on T6 (67.4\%) and T5 (76.7\%) reflect pathologies with distinct auscultatory signatures that map well to clinical descriptions. Performance on T3 (51.3\%) and T9 (52.1\%) remains near chance, reflecting the difficulty of distinguishing subtle cough variations (T3) and grading severity across five classes from text alone (T9).

\begin{table}[t]
    \centering
    \scriptsize
    \caption{Ablation results (mean \% AUC across 9 tasks).}
    \label{tab:ablation}
    \vspace{-3pt}
    \begin{tabular}{l ccc}
        \toprule
        Configuration & Lin.\ Probe $\uparrow$ & $k$\text{-}NN $\uparrow$ & Zero Shot $\uparrow$ \\
        \midrule
        Audio Encoder from Scratch    & 66.0 & 58.8 & 55.1 \\
        Random Negatives      & 70.6 & 63.9 & 53.5 \\
        No $\mathcal{L}_{\text{MSE}}$ & 71.1 & 62.8 & 54.1 \\
        Frozen Audio Backbone & 66.9 & 58.5 & 48.8 \\
        BERT Text Encoder & 65.1 & 59.6 & 49.9 \\
        Negative Sampling K=100 & 69.9 & 61.1 & 54.2 \\
        \midrule
        \textbf{REACH (Ours)}  & \textbf{71.6} & \textbf{65.2} & \textbf{61.3} \\
        \bottomrule
    \end{tabular}
    \vspace{-0.6cm}
\end{table}

\vspace{-0.15cm}
\subsection{Ablation Studies}
\label{sec:ablation}

Table~\ref{tab:ablation} isolates each component's contribution. A consistent pattern emerges: several ablations maintain linear probing while severely degrading zero-shot AUC, indicating the removed component is essential for cross-modal alignment, not unimodal feature quality.

Training the audio encoder from scratch yields near-random zero-shot AUC (55.1) despite adequate linear probing (66.0), confirming that alignment requires an established acoustic manifold to reorganize. Replacing FAISS-based distant negatives with random in-batch sampling preserves linear probing (70.6) but collapses zero-shot AUC to 53.5, as clinical reports for related pathologies are near-paraphrases that only distant negatives can disambiguate. Removing $\mathcal{L}_{\text{MSE}}$ produces the most revealing dissociation: a 7.2-point zero-shot drop (54.1) with a only 0.5-point linear probing decrease, showing that without reconstruction regularization, the contrastive objective distorts pre-trained geometry into a space no longer coherent for text anchor mapping. Freezing the audio backbone and training only projection heads yields the weakest zero-shot AUC (48.8), confirming that partial encoder adaptation is necessary. Substituting MedSigLIP with BERT drops zero-shot AUC by 11.4 points (49.9\%), confirming general-purpose text encoders lack clinical grounding for effective alignment. Increasing negative sampling to $K{=}100$ also degrades zero-shot AUC (54.2\%), indicating similar negatives fail to provide clear boundaries between unrelated pathologies.

\section{Conclusion}
We introduced REACH, a framework for aligning pre-trained respiratory audio encoders with clinical text representations, enabling zero-shot classification without paired audio-report data. Our results suggest that the primary barrier to zero-shot respiratory diagnostics is not acoustic representation quality, which existing self-supervised encoders already capture well, but rather their disconnection from clinical semantics. By treating an off-the-shelf LLM as a metadata-to-report translator, even discrete categorical annotations can be elevated into rich linguistic anchors sufficient for contrastive alignment. Most notably, our framework surpasses baselines trained on 57\% more data and models with orders-of-magnitude more parameters, suggesting that targeted semantic bridging of mature unimodal models is a more sample-efficient paradigm than scaling pre-training or building multimodal architectures from scratch. We believe this principle generalizes beyond respiratory audio: any clinical domain with capable unimodal encoders and structured metadata is a candidate for the same alignment strategy.

\newpage

\section{Generative AI Use Disclosure}
\small{Generative AI tools were used solely for language editing and polishing to improve clarity and readability of the manuscript. All technical content, experimental design, analysis, and conclusions were created by the authors. The authors take full responsibility for the content of this paper.}

\section{Acknowledgments}
\small{This work was supported by the NWO AiNed Fellowship Grant of A.S., and in part by Google.org and the Google Cloud Research Credits program through the Gemini Academic Program. We also acknowledge the use of the Dutch National Supercomputer Snellius for essential computational tasks.}

\bibliographystyle{IEEEtran}
\bibliography{mybib}

\end{document}